\def\year{2022}\relax
\documentclass[letterpaper]{article}
\pdfoutput=1
\usepackage{aaai22}  
\usepackage{times}  
\usepackage{helvet}  
\usepackage{courier}  
\usepackage[hyphens]{url}  
\usepackage{graphicx} 
\usepackage{natbib}  
\usepackage{caption} 
\DeclareCaptionStyle{ruled}{labelfont=normalfont,labelsep=colon,strut=off} 
\usepackage{amssymb}
\usepackage{newfloat}
\usepackage{listings}
\usepackage{booktabs} 
\usepackage{paralist}
\usepackage[inline]{enumitem}
\usepackage{multirow}
\usepackage{subfigure}
\usepackage[np, autolanguage]{numprint}
\usepackage{xcolor}
\usepackage[noend]{algpseudocode}
\usepackage[skip=1pt]{caption}
\usepackage{acronym}
\usepackage{graphicx}
\usepackage{amsmath}

\newcommand{\OurUniversity}{the University of Amsterdam}
\acrodef{UvA}{University of Amsterdam}
\acrodef{AI}{Artificial Intelligence}
\acrodef{FACT}{Fairness, Accountability, Confidentiality and Transparency}
\acrodef{FACT-AI}{Fairness, Accountability, Confidentiality and Transparency in Artificial Intelligence}
\acrodef{MLRC}{Machine Learning Reproducibility Challenge}

\title{Reproducibility as a Mechanism for Teaching Fairness, Accountability, Confidentiality, and Transparency in Artificial Intelligence}

\author {
    Ana Lucic,\equalcontrib{}\textsuperscript{\rm 1}
    Maurits Bleeker,\equalcontrib{}\textsuperscript{\rm 1}
    Sami Jullien,\textsuperscript{\rm 2}
    Samarth Bhargav,\textsuperscript{\rm 1}
    Maarten de Rijke\textsuperscript{\rm 1}
}
\affiliations{
   \textsuperscript{\rm 1}University of Amsterdam\qquad \textsuperscript{\rm 2}AIRLab, University of Amsterdam\\
   \{a.lucic, m.j.r.bleeker, s.jullien, s.bhargav, m.derijke\}@uva.nl
}

\begin{document}

\maketitle

\begin{abstract}
In this work, we explain the setup for a technical, graduate-level course on Fairness, Accountability, Confidentiality, and Transparency in Artificial Intelligence (FACT-AI) at \OurUniversity{}, which teaches FACT-AI concepts through the lens of reproducibility. 
The focal point of the course is a group project based on reproducing existing FACT-AI algorithms from top AI conferences and writing a corresponding report. 
In the first iteration of the course, we created an open source repository with the code implementations from the group projects. 
In the second iteration, we encouraged students to submit their group projects to the Machine Learning Reproducibility Challenge, resulting in 9 reports from our course being accepted for publication in the ReScience journal. 
We reflect on our experience teaching the course over two years, where one year coincided with a global pandemic, and propose guidelines for teaching FACT-AI through reproducibility in graduate-level AI study programs. 
We hope this can be a useful resource for instructors who want to set up similar courses in the future. 
\end{abstract}

\section{Introduction}
\label{section:introduction}

For several decades, \OurUniversity{} has offered a research-oriented Master of Science (MSc) program in \acf{AI}. 
The main focus of the program is on the technical, machine learning (ML) aspects of the major sub-fields of \ac{AI}, such as computer vision, information retrieval, natural language processing, and reinforcement learning.
One of the most recent additions to the MSc AI curriculum is a mandatory course on \ac{FACT-AI}. 
This course was first taught during the 2019--2020 academic year and focuses on teaching FACT-AI topics through the lens of reproducibility. 
The main project involves students working in groups to re-implement existing FACT-AI algorithms from papers in top AI venues. 
There are approximately 150 students enrolled in the course each year. 

The motivation for the course came from the MSc AI students themselves, who often play an important role in shaping the curriculum in order to meet the evolving requirements of researchers in both academia and industry. 
As the influence of \ac{AI} on decision making is becoming increasingly prevalent in day-to-day life, there is a growing consensus that stakeholders who take part in the design or implementation of \ac{AI} algorithms should reflect on the ethical ramifications of their work, including developers and researchers \cite{campbell2021_responsible}. 
This is especially true in situations where data-driven AI systems affect some demographic sub-groups differently than others \cite{angwin2016machine, o2017ivory}.
As a result, our students have shown an increased interest in the ethical issues surrounding AI systems and requested that the university put together a new course focusing on responsible AI.

Since our MSc AI program is characterized by a strong emphasis on understanding, developing, and building AI algorithms, we believe that a new course on responsible AI in this program should also have a hands-on approach.
The course is designed to address technical aspects of key areas in responsible AI:
\begin{enumerate*}[label=(\roman*)]
\item fairness, 
\item accountability, 
\item confidentiality, and 
\item transparency, 
\end{enumerate*}
which we operationalize through a reproducibility project. 
We believe a strong emphasis on reproducibility is important from both an educational point of view and from the point of view of the AI community, since the (lack of) reproducible results has become a major point of critique in \ac{AI} \cite{hutson2018artificial}. 
Moreover, the starting point of almost any junior AI researcher (and most AI research projects in general) is re-implementing existing methods as baselines. 
The FACT-AI course is situated at a point in the program where students have learned the basics of ML and are ready to start experimenting with, and building on top of, state-of-the-art algorithms. 
Given that our MSc AI program is fairly research-oriented, it is important for students to experience the process of reproducing work done by others (and how difficult this is) at an early stage in their careers. 
We also believe reproducibility is a fundamental component of FACT-AI: the cornerstone of fair, accountable, confidential and transparent \ac{AI} systems is having correct and reproducible results. 
Without reproducibility, it is unclear how to judge if a decision-making algorithm adheres to any of the FACT principles. 

In the 2019--2020 academic year, we operationalized our learning ambitions regarding reproducibility by publishing a public repository with selected code implementations and corresponding reports from the group projects.
In the 2020--2021 academic year, we took the projects one step further and encouraged students to submit to the \ac{MLRC},\footnote{\url{https://paperswithcode.com/rc2020}} a competition that solicits reproducibility reports for papers published in conferences such as NeurIPS, ICML, ICLR, ACL, EMNLP, CVPR and ECCV. 
Although the MLRC broadly focuses on all papers submitted to these conferences, we focus exclusively on papers covering FACT-AI topics in our course. 
Submitting to the MLRC gives students a chance to experience the whole AI research pipeline, from running experiments, to writing rebuttals, to receiving the official notifications. 
Of the 23 papers that were accepted to the MLRC in 2021, 9 came from groups in the FACT-AI course at \OurUniversity{}. 

In this work, we describe the \textit{Fairness, Accountability, Confidentiality, and Transparency in Artificial Intelligence} course: a one month, full-time course based on examining ethical issues in AI using reproducibility as a pedagogical tool. 
Students work in groups to re-implement (and possibly extend) existing algorithms from top AI venues on \ac{FACT-AI} topics. 
The course also includes lectures that cover the high-level principles of FACT-AI topics, as well as paper discussion sessions where students read and digest prominent FACT-AI papers. 
In this paper, we outline the setup for the \ac{FACT-AI} course and the experiences we had while running the course during the 2019--2020 and 2020--2021 academic years at \OurUniversity{}. 

The remainder of this paper is structured as follows. 
In Section~\ref{section:related-work} we discuss related work, specifically other courses about responsible AI. 
In Section~\ref{section:reproducibility}, we detail ongoing reproducibility efforts in the AI community. 
In Section~\ref{section:learning-objectives}, we explain the learning objectives for our course, and explain how we realized those objectives in Section~\ref{section:coursesetup}. 
We reflect on the feedback we received about the course in Section~\ref{section:feedback}, as well as what worked (Section~\ref{section:whatworked}) and what did not (Section~\ref{section:whatdidnt}), before concluding in Section~\ref{section:conclusion}.

\section{Related Work}
\label{section:related-work}

There have been multiple calls for introducing ethics in computer science courses in general, and in AI programs in particular \cite{angwin2016machine, leonelli2016locating, o2017ivory,national2018data, singer2018tech, skirpan2018ethics, grosz2019embedded, salz2019integrating, danyluk2021computing}. 
Several surveys have investigated how existing responsible computing courses are organized \cite{peck2017,fiesler2020we, garret2020morethan, raji_you_2021}. 

\subsection{Characterizing Responsible AI Courses}
There are two primary approaches to integrating such components into the curriculum: \begin{enumerate*}[label=(\roman*)]
\item stand-alone courses that focus on ethical issues such as FACT-AI topics, and 
\item a holistic curriculum where ethical issues are introduced and tackled in each course \cite{peck2017, fiesler2020we, garret2020morethan}. 
\end{enumerate*}
In general, the latter is rare \cite{peck2017, salz2019integrating, fiesler2020we}, and can be difficult to organize due to a lack of qualified faculty or relevant expertise \cite{bates2020integrating, raji_you_2021}. 
We opt for the first approach since our course is a new addition to an existing study program. 

\citet{fiesler2020we} analyze 202 courses on ``tech ethics''. 
Their survey examines 
\begin{enumerate*}[label=(\roman*)]
\item the departments the courses are taught from, as well as the home departments of the course instructors, 
\item the topics covered in the courses, organized into 15 categories, and 
\item the learning outcomes in the courses. 
\end{enumerate*}
In our case, both the FACT-AI course and its instructors are from the Informatics Institute of the Faculty of Science at \OurUniversity{}. Our learning objectives (see Section~\ref{section:learning-objectives}) correspond to the following learning objectives from \citet{fiesler2020we}: ``Critique'', ``Spot Issues'', and ``Create solutions''. 
According to their content topic categorization, our course includes ``AI \& Algorithms'' and ``Research Ethics'': the former since the course deals explicitly with \ac{AI} algorithms under the FACT-AI umbrella, and the latter due to its focus on reproducibility. 
We note that ``AI \& Algorithms'' is only the fifth-most popular topic according to the survey, after ``Law \& Policy'', ``Privacy \& Surveillance'', ``Philosophy'', and ``Inequality, Justice \& Human Rights'' \citep[see Table 2,][]{fiesler2020we}). 
Although we believe these topics are important, we also wanted to avoid the feeling that the course was a ``distraction from the real material'' \cite{lewis2021teaching}, especially since 
\begin{enumerate*}[label=(\roman*)]
\item the majority of our students are coming from a technical background into a technical MSc program, and 
\item the FACT-AI course is mandatory for all students in the MSc AI program. 
\end{enumerate*}

\subsection{Similar Responsible AI Courses}
The two courses that are the most similar to ours are those of \citet{lewis2021teaching} and \citet{yildiz2021reproducedpapers}. 

\citet{lewis2021teaching} describe a course for responsible data science. 
Similar to our course, they focus on the technical aspects of AI, involving lectures, readings, and a final project. 
However, their course differs from ours since the main project in their course is focused on examining the interpretability of an automated decision making system, while the main project in our course is focused on reproducibility. 

\citet{yildiz2021reproducedpapers} describe a course based on reproducing experiments from AI papers, focusing on ``low-barrier'' reproducibility. 
Similar to our course, this course involves replicating a paper from scratch or reproducing the experiments using existing code, performing hyperparameter sweeps, and testing with new data or with variant algorithms. 
Another similarity is that they released a public repository of re-implemented algorithms,\footnote{\url{https://reproducedpapers.org/}} which we also did for the first iteration of our course.\footnote{\url{https://github.com/uva-fact-ai-course/uva-fact-ai-course}} 
However, their course differs from ours since theirs focuses on AI papers in general, while our course focuses exclusively on FACT-AI papers. 

There are several courses that focus more on the philosophical or social science perspectives of AI ethics.
\citet{green2021aiethics} describes an undergraduate AI ethics course that teaches computer science majors to analyse issues using different ethical approaches and how to incorporate these into an \textit{explicit} ethical agent. 
\citet{shenvalue2021} introduce a toolkit in the form of ``Value Cards'' to inform students and practitioners about the social impacts of ML models through deliberation.
\citet{green2020argument} propose an approach to ethics education using ``argument schemes'' that summarize key ethical considerations for specialized domains such as healthcare or national defense.
\citet{Furey2018IntroducingET} introduce ethics concepts, primarily utilitarianism, into an existing AI course about autonomous vehicles by studying several variations of the Trolley Problem.  
\citet{burton2018} teach ethics through science fiction stories complemented with philosophy papers, allowing students to reflect and debate difficult content without emotional or personal investment since the stories are not tied to ``real'' issues. 
\citet{skirpan2018ethics} describe an undergraduate course on human-centred computing which integrates ethical thinking throughout the design of computational systems. 
Unlike these courses, our course focuses more on the technical aspects of ethical AI.  
However, incorporating such non-technical perspectives is something we would like to do in future iterations of our course, perhaps through one of the mechanisms employed by some of these courses. 

\section{Integrating Reproducibility of AI Research into the FACT-AI Course}
\label{section:reproducibility}

There have been several criticisms about the lack of reproducibility in AI research. 
Some have postulated this is due to a combination of unpublished code and high sensitivity of training parameters \citep{hutson2018artificial}, while others believe the rapid rate of progress in ML research results in a lack of empirical rigor \citep{sculley2018winner}. 
Although typically well-intentioned, some papers may disguise speculation as explanation, obfuscate content behind math or language, and fail to attribute the correct sources of empirical gains \citep{lipton2019research}. 

Several efforts have been made to investigate and increase the reproducibility of AI research. 
In 2021, NeurIPS introduced a paper checklist including questions about reproducibility, along with a template for submitting source code as supplementary material \citep{neuripschecklist}. 
The Association of Computing Machinery introduced a badging system that indicates how reproducible a paper is \citep{acm-artifact}. 
Papers with Code is an organization that provides links to official code repositories in arXiv papers \citep{paperswithcode2020}. 
It also hosts an annual Machine Learning Reproducibility Challenge (MLRC): a community-wide effort to investigate the reproducibility of papers accepted at top AI conferences, which we incorporated into the 2020--2021 iteration of the FACT-AI course. 

In an ideal scenario, reproducibility issues would be handled prior to publication \citep{sculley2018winner}, 
but it can be difficult to catch such shortcomings in the review process due to the increasing number of papers submitted to AI conferences. 
Therefore, we believe it is of utmost importance that the next generation of AI researchers -- including our own students -- can 
\begin{enumerate*}[label=(\roman*)]
\item identify and 
\item avoid these pitfalls while conducting their own work. 
\end{enumerate*}
This, in combination with the fact that reproducibility is a fundamental component of responsible AI research, is why we opted to teach the FACT-AI course through the lens of reproducibility. 

Our course is centered around a group project where students re-implement a recent FACT-AI algorithm from a top AI conference. 
This project has three components:
\begin{enumerate*}[label=(\roman*)]
\item a reproducibility report, 
\item an associated code base, and 
\item a group presentation. 
\end{enumerate*}
In Section~\ref{section:assignment}, we provide more details on the project and the outputs it resulted in.

\section{Learning Objectives}
\label{section:learning-objectives}

In the FACT-AI course, we aim to make students aware of two types of responsibility: (i) towards society in terms of potential implications of their research, and (ii) towards the research community in terms of producing reproducible research. 
In this section, we outline the learning objectives for the FACT-AI course and explain how it fits within the context of the MSc AI program at \OurUniversity{}. 

Table~\ref{tab:msc_program} shows the setup of the first year of the 2-year MSc AI program. Each semester at \OurUniversity{} is divided into three periods: two 8-week periods followed by one 4-week period. During an 8-week period, students follow two courses in parallel. During the 4-week period, they only follow a single course. 
The FACT-AI course takes place during the 4-week period at the end of the first semester, after students have taken Computer Vision 1, Machine Learning 1, Natural Language Processing 1 and Deep Learning 1. 
It is the only course students follow during this period, so we believe it is beneficial to have them focus on one main project -- reproducing an existing FACT-AI paper. 
The learning objectives for the course are as follows: 

\begin{itemize}
\setlength\itemsep{1em}
	\item \textbf{LO \#1: Understanding FACT topics.} Students can explain the major notions of fairness, accountability, confidentiality, and transparency that have been proposed in the literature, along with their strengths and weaknesses.
	\item \textbf{LO \#2: Understanding algorithmic harm.} Students can explain, motivate, and distinguish the main types of algorithmic harm, both in general and in terms of concrete examples where AI is being applied.
	\item \textbf{LO \#3: Familiarity with FACT methods.} Students are familiar with recent peer-reviewed algorithmic approaches to fairness, accountability, confidentiality, and transparency in the literature. 
    \item \textbf{LO \#4: Reproducing FACT solutions.} Students can assess the degree to which recent algorithmic solutions are effective, especially with respect to the claims made in the original papers, while understanding their limitations and shortcomings. 
\end{itemize}

\section{Course Setup}
\label{section:coursesetup}
The FACT-AI course is organized around 
\begin{enumerate*}[label=(\roman*)]
    \item lectures, 
    \item paper discussions, and
    \item a group project. 
\end{enumerate*}
It has had two iterations so far: the 2019--2020 iteration was taught in person, while the 2020--2021 iteration was taught online due to the COVID-19 pandemic. 
In this section, we detail how we realized the learning objectives from Section~\ref{section:learning-objectives} and describe the challenges in adapting the course to an online format.

\subsection{Lectures}
To further the understanding of FACT-AI topics (LO1), we provide one general lecture for each of the 4 topics, along with a lecture specifically about reproducibility.
Lectures are an opportunity for students to familiarize themselves with algorithmic harm (LO2). Students are encouraged to ask questions that lead to discussions about potential harm done by applications of AI. 
This was more challenging in the second iteration of the course due to the online format, but we hope that facilitating such discussions will be more straightforward once we return to in-person classes. 

In addition to the general lectures, we also include some guest lectures. 
These are used to either discuss specific types of algorithmic harm (LO2), examine specific FACT-AI algorithms in depth (LO3), or expand on the non-technical aspects of FACT-AI. 
Some examples of guest lectures include a lecture on AI accountability from a legal perspective by an instructor from the law department of \OurUniversity{}, and a lecture by two former FACT-AI students who explained how they turned their group project into an ICML 2021 workshop paper \cite{neely2021order}. 

\subsection{Paper Discussions}
The goal of the paper discussion sessions is for students to learn about prominent FACT-AI methods (LO3), and learn to think critically about the claims made in the papers we discuss (LO4). 
Students first read a seminal FACT-AI paper on their own while trying to answer the following questions: 
\begin{itemize}[leftmargin=*,nosep]
\item What are the main claims of the paper?
\item What are the research questions?
\item Does the experimental setup make sense, given the research questions?
\item What are the answers to the research questions? Are these supported by experimental evidence?
\end{itemize}

\noindent%
Once students have read the papers, they participate in smaller discussion sessions with their peers about their answers to the questions above. 
After each discussion session, all the groups are brought back together for a ``dissection'' session, where an instructor goes over the same seminal paper, giving an overview of the papers' strengths and weaknesses. 

Each session was presented by a different instructor to show that there is no single way of examining a research paper, and that different researchers will bring different perspectives to their assessment of papers. 
The following papers were covered during the discussion sessions: 
\citet{hardt_equality_2016} on fairness; 
\citet{ribeiro_why_2016} on transparency; and 
\citet{Abadi_2016} on confidentiality.

\begin{table}[t]
\caption{The first year of the MSc AI program at \OurUniversity{}.}
\label{tab:msc_program}
\centering
\footnotesize
\setlength{\tabcolsep}{3pt}
\begin{tabular}{l c@{~}c@{~}c c@{~}c@{~}c c}
\toprule
Course & \multicolumn{3}{c}{Sem.\ 1} & \multicolumn{3}{c}{Sem.\ 2} & EC \\
\midrule
     Computer Vision 1 & $\blacksquare$ & $\square$ & $\square$ & $\square$ & $\square$ & $\square$ & 6 \\
     Machine Learning 1 & $\blacksquare$ & $\square$ & $\square$ & $\square$ & $\square$ & $\square$ & 6 \\
     Natural Language Processing 1 & $\square$ & $\blacksquare$ & $\square$ & $\square$ & $\square$ & $\square$ & 6 \\
     Deep Learning 1 & $\square$ & $\blacksquare$ & $\square$ & $\square$ & $\square$ & $\square$ & 6\\
     Fairness, Accountability, Confidentiality  & $\square$ & $\square$ & $\blacksquare$ & $\square$ & $\square$ & $\square$ & 6 \\
     and Transparency in AI \\
     Information Retrieval 1 & $\square$ & $\square$ & $\square$ & $\blacksquare$ & $\square$ & $\square$ & 6 \\
     Knowledge Representation and Reasoning & $\square$ & $\square$ & $\square$ & $\blacksquare$ & $\square$ & $\square$ & 6 \\
     Elective 1 & $\square$ & $\square$ & $\square$ & $\square$ & $\blacksquare$ & $\square$ & 6 \\
     Elective 2 & $\square$ & $\square$ & $\square$ & $\square$ & $\blacksquare$ & $\square$ & 6 \\
     Elective 3 & $\square$ & $\square$ & $\square$ & $\square$ & $\square$ & $\blacksquare$ & 6 \\
\bottomrule
\end{tabular}
\end{table}

\subsection{Group Project}
\label{section:assignment}

\subsubsection{Reproduction of a FACT-AI paper.}
The purpose of the group project is to have students investigate the claims made by the authors of recent FACT-AI papers by diving into the details of the methods and their implementations.  
Using what they have learned from the paper discussion sessions, students work in groups to re-implement an existing FACT-AI algorithm from a top AI conference and re-run the experiments in the paper to determine the degree to which they are reproducible (LO4). 
If the code is already available, then they must extend the method in some way. 
The project consisted of three deliverables: 
\begin{enumerate*}[label=(\roman*)]
\item a reproducibility report, 
\item an associated code base, and 
\item a group presentation. 
\end{enumerate*}

In order to ensure the project is feasible, we select 10--15 papers in advance for groups to choose from. 
Our criteria for including papers is as follows:
\begin{itemize}[leftmargin=*,nosep]
    \item The paper is on a FACT-AI topic. 
    \item At least one dataset in the paper is publicly available.
    \item Experiments can be run on a single GPU (which we provide access to).
    \item It is reasonable for a group of 3--4 MSc AI students to re-implement the paper within the timeframe of the course. In our case, students work on this project for one-month full-time. 
\end{itemize}

\noindent%
To ease the load for our teaching assistants (TAs), we have several groups working on the same paper. 
We assign papers to TAs based on their interests by asking them to rank the set of candidate papers in advance. 
We also encourage them to suggest alternative papers provided they fit the criteria. 
The TAs read the papers before the course starts in order to ensure they have a sufficient, in-depth understanding of the work such that they can guide students through the project. 
This also serves as an extra feasibility check, to ensure that the papers are indeed a good fit for our course.

Each group writes a report about their efforts following the structure of a standard research paper (i.e., introduction, methodology, experiments, results, conclusion). 
They also include aspects specific to reproducibility such as explaining the difficulties of implementing certain components, as well as describing any communication they had with the original authors.
In addition to the source code, students provide all results in a Jupyter notebook along with a file to install the required environment. 

\subsubsection{First Iteration: Contributing to an Open Source Repository.}
In the 2019--2020 iteration of the course, we created a public repository on GitHub, which contains a selection of the implementations done by our students: \url{https://github.com/uva-fact-ai-course/uva-fact-ai-course}. 
The TAs who assisted with the course decided which implementations to include and cleaned up the code so it all fit into one cohesive repository. 
This had multiple motivations.
First, it taught students how to improve the reproducibility of their own work by releasing the code, while also giving them a sense of contributing to the open-source community. 
Second, a public repository can serve as a starting point for personal development in their future careers; companies often ask to see existing code or projects that prospective employees have worked on. 
Some students added the project to their CVs, while others wrote blog posts about their experiences,\footnote{https://omarelb.github.io/self-explaining-neural-networks/}  linking back to the repository.

\subsubsection{Second Iteration: The Machine Learning Reproducibility Challenge.}
In the 2020--2021 iteration of the course, we formally participated in the annual MLRC run by Papers with Code \citep{paperswithcode2020} in order to expose our students to the peer-review process. 
This gave students something to strive towards and offered perspectives beyond simply getting a grade for the project. 
Most importantly, it gave them the opportunity to experience the full research pipeline: 
\begin{enumerate*}[label=(\roman*)]
	\item reading a technical paper, 
	\item understanding a paper's strength and weaknesses, 
	\item implementing (and perhaps also extending) the paper, 
	\item writing up the findings, 
	\item submitting to a venue with a deadline, 
	\item obtaining feedback, 
	\item writing a rebuttal, and
	\item receiving the official notification. 
\end{enumerate*}
To encourage students to formally submit to the MLRC, we offered a 5\% boost to their final grades if they submitted. 
Of the 32 groups in the FACT-AI course, 30 (94\%) groups submitted their reproducibility reports to the MLRC, of which 9 groups had their papers accepted.

\subsection{Taking the Course Online}
The second iteration of the course was taught in January 2021, when the COVID-19 pandemic forced us to move classes and interactions online. 
Students made use of various channels to communicate: WhatsApp, Discord, and Slack.  
Canvas was the primary mode of communication between the instructors and the students, allowing students to ask questions and instructors to communicate various announcements. 

Lectures were live, with frequent Q\&A breaks to stimulate interactivity. 
Paper discussion sections were organized in different online meeting subrooms where students discussed the papers together.  
This proved to be a challenge: while some subrooms had productive discussions, others struggled to get the conversation going. 

The reproducibility project was more difficult to launch remotely. 
Since students had done online classes for their entire first semester, some struggled to find a group of fellow students to team up with, especially those coming from outside the MSc AI program. 
Overall, while we had various communication means, the lack of physical interaction due to COVID-19 slowed down our course organization.

\section{Feedback}
\label{section:feedback}

In this section, we discuss the feedback we received about the course from the perspective of participating students (Section~\ref{section:feedback-students}) and from the MLRC reviews (Section~\ref{section:feedback-mlrc}). 
 
\subsection{Feedback from Students}
\label{section:feedback-students}
Both iterations of the course were evaluated using the standard evaluation procedure provided by \OurUniversity. 
However, only 16\% of students filled out the evaluation form (23 out of 144) in the 2020--2021 iteration, potentially because the evaluation forms were administered online instead of in-person. 
According to the evaluation procedure at our university, this is not enough for a reliable quantitative estimate of student satisfaction.
Therefore, we focus on the 2019--2020 iteration when reporting student satisfaction statistics, since 53\% of students filled out the form (79 out of 149) that year. 

The vast majority of students were (very) satisfied with the course overall (67.8\%). 
More specifically, students were (very) satisfied with the following dimensions: 
\begin{itemize}[leftmargin=*,nosep]
    \item Academic challenge (75.2\%)
    \item Supervision (76.9\%)
    \item Feedback (81.3\%)
    \item Workload (91.3\%)
    \item Level of the course (79.7\%)
    \item Level of the report (94.8\%)
    \item Level of the presentation (96.6\%)
\end{itemize}
Table~\ref{tab:feedback}(a) shows some of the qualitative feedback we received from students. 
Based on this, we believe these high scores are mostly the result of the reproducibility project. 
Students enjoyed doing the project, especially due to the intensive supervision from our experienced TAs. 
The dimensions where we received the lowest scores were on the lectures and the final presentation, where only 30.6\% and 30.2\% were (very) satisfied with these aspects, respectively. 
This may be because we only provided four (high-level) lectures on each of the four topics, in order to give students as much time as possible to focus on the reproducibility project. 
However, it should be noted that the overall scores for these components were not poor, but average: 3.1/5 for lectures and 3.0/5 for the presentation. 

\subsection{Feedback from the MLRC}
\label{section:feedback-mlrc}

Of the 30 reproducibility reports submitted to the MLRC in the 2020--2021 iteration, 9 were accepted for publication in the ReScience Journal.  
In total, the MLRC accepted 23 reports, meaning that almost 40\% of the reports accepted to the MLRC were from \OurUniversity{}.\footnote{https://openreview.net/group?id=ML\_Reproducibility\_Challenge/2020}
 
The reviews were mostly positive, with the general consensus being that most teams had gone beyond the general expectation of simply re-implementing the algorithm and re-running the experiments. 
Our TAs encouraged students to examine the generalizability of the work that was reproduced, either by trying new datasets or hyperparameters, or by performing ablation studies. 
Multiple reproducibility reports managed to question the results of the original papers with experimentally-supported claims. 
Importantly, some reviewers emphasized that these reproducibility studies were solid starting points for future research projects. 
For the reports that were rejected, the main critiques were that
\begin{enumerate*}[label=(\roman*)]
    \item only a fraction of the original work was reproduced, or 
    \item no new insights were given.
\end{enumerate*}
Some projects also had flaws in the experimental setup. 
See Table~\ref{tab:feedback}(b) and  ~\ref{tab:feedback}(c) for quotes from the MLRC reviews.

\section{What Worked}
\label{section:whatworked}

Understanding and re-implementing the work of other researchers is not a trivial task, especially for first-year MSc students. There were several aspects of the setup that we believe were beneficial for the students, which we organize along three dimensions: 
\begin{enumerate*}[label=(\roman*)]
    \item general, 
    \item concerning FACT-AI, and 
    \item concerning reproducibility.
\end{enumerate*}
We believe each of these factors are important for a successful implementation of this course, or other similar courses. 

\subsection{General}

\subsubsection{Timing of the course.}
It is important that students have prior knowledge of ML theory as well as some programming experience before completing a project-based course in groups. At \OurUniversity{}, the FACT-AI course takes place after students have completed 4 ML-focused courses (see Table~\ref{tab:msc_program}). 
We believe it is extremely important that students have access to adequate preparation, especially in terms of programming experience, before setting off to reproduce experiments from prominent AI papers. 
Without this prior knowledge, we believe such a project would not be feasible in the allotted time frame. 

\subsubsection{Regular contact with experienced TAs.}
The TAs are there to help with two main components: 
\begin{enumerate*}[label=(\roman*)]
\item understanding the paper, and 
\item debugging the implementation process. 
\end{enumerate*}
In practice, we found that it is extremely important for the TAs to have excellent programming experience since this is the main aspect students need help with. 
We also had a dedicated Slack workspace for the TAs and course instructors to keep in touch regularly. 

Since our course is only four weeks long, we found it was important for students to have regular contact with their TAs to ensure no one got stuck in the process. 
For the first (in-person) iteration of the course, groups had one-hour tutorials with their TAs twice per week, where all groups that were working on the same paper (and therefore had the same TA) were in the same tutorial. 
Since they were all working on the same paper, there were many overlapping questions, and students found it beneficial to be able to share their experiences with one another. 
For the second (online) iteration of the course, we thought it would be challenging to ensure each group got the attention they needed if everyone was in one large online tutorial, so the TAs met with each group separately for 30 minutes, twice per week. 

\subsubsection{Early feedback on the reports.}
Approximately halfway through the course, we asked students to submit a draft report to their TAs in order to get feedback. We found this significantly increased the quality of the final reports. 

\subsection{Concerning FACT-AI}

\subsubsection{Emphasizing the technical perspective of FACT-AI.}
Given that the FACT-AI course is situated in the context of a technical, research-oriented MSc, having students re-implement research papers from top AI conferences was an effective way to teach FACT-AI topics for our students. 
Teaching FACT-AI from a primarily technical perspective 
aligns well with what students expect from the MSc AI program at \OurUniversity{}.
Although we believe a technical focus makes sense for our MSc program, we also believe it is important to incorporate non-technical perspectives into the course -- see Section~\ref{section:non-technical}. 

\subsubsection{Creating resources for the FACT-AI community.}
We believe a significant motivating factor for students was creating concrete output that extended beyond simply completing a project for a course: creating resources for the FACT community. 
In the 2019--2020 iteration, this was done by creating a public repository with the best FACT-AI algorithm implementations, as selected by the TAs. 
In the 2020--2021 iteration, this was done by publicly submitting their reproducibility reports about FACT-AI algorithms to the MLRC, where the accepted reports were published in the ReScience Journal. 
In the future, we plan to continue aligning our course with the MLRC since we found the process extremely beneficial for our students.

\subsection{Concerning Reproducibility}

\subsubsection{Including extension as part of reproducibility.}
If source code was already available for the paper -- which is fortunately becoming increasingly common for AI research papers -- we asked students to think about how to extend the paper since the implementation was already available. This resulted in some creative and interesting ideas in the reports, and we believe this is why our students performed well at the MLRC (see Section~\ref{section:feedback-mlrc})

\subsubsection{Simple grading setup.}
For a 4-week, project-based course, we found it was beneficial for students to focus one main deliverable consisting of three components: 
\begin{enumerate*}[label=(\roman*)]
\item the reproducibility report, 
\item the associated code base, and
\item the group presentation. 
\end{enumerate*}
The report that students submitted for the course was the same one they submitted to the MLRC. This way, participating in the MLRC was not an extra task but rather an integral part of the course.

\section{What Could Be Improved}
\label{section:whatdidnt}

Although we believe both iterations of the course went well, there are several aspects of the setup that we believe could use some improvement and other instructors should consider if they plan to implement a similar course. 

\subsection{General}

Given that this is the first time most students are formally submitting a paper, it is not surprising that there were some logistical issues. 
Some groups made minor mistakes such as forgetting to submit their work double-blind or slightly missing the submission deadline. 
We also had some groups who wrote the introduction sections of their papers as an introduction to the FACT-AI course, rather than an introduction to the topic they were working on. 
In future iterations, we will explicitly state the standard procedures of writing and submitting a paper and provide some examples. 

\subsection{Concerning FACT-AI}
\label{section:non-technical}

Although focusing primarily on the technical aspects of FACT-AI is an effective way to engage our technical students in socially-relevant AI problems, we also believe that they would benefit from additional non-technical perspectives on FACT-AI topics. 
In the future, we plan to include perspective from outside of computer science through 
\begin{enumerate*}[label=(\roman*)]
\item additional guest lectures, 
\item workshop sessions \citep{skirpan2018ethics,shenvalue2021}, and 
\item broader impact statements \citep{campbell2021_responsible} in the reproducibility reports. 
\end{enumerate*}

\subsection{Concerning Reproducibility}
In future iterations, we believe it would be useful to show students more examples of what a high-quality reproducibility paper looks like and explain in-depth why it is high-quality. 
These could be papers that were previously accepted to the MLRC, or papers from other reproducibility efforts outlined in Section~\ref{section:reproducibility}. 
We want the students to understand what makes a paper a good (reproducibility) paper, that is, it has a set of (reproducibility) claims, it argues for these claims, and shows evidence to support these claims.

\section{Conclusion}
\label{section:conclusion}

In this paper, we share our setup for the FACT-AI course at \OurUniversity{}, which teaches FACT-AI topics through reproducibility. 
The course set out to give students 
\begin{enumerate*}[label=(\roman*)]
    \item an understanding of FACT-AI topics, 
    \item an understanding of algorithmic harm, 
    \item familiarity with recent FACT-AI methods, and
    \item an opportunity to reproduce FACT-AI solutions, 
\end{enumerate*}
through a combination of lectures, paper discussion sessions and a reproducibility project. 

Through their projects, our students engaged with the open-source community by creating a public code repository (in the 2019--2020 iteration), as well as with the research community via successful submissions to the MLRC challenge (in the 2020--2021 iteration). 
We also detail how the 2020--2021 iteration brought about its own unique set of challenges due to the COVID-19 pandemic. 

In this course, we illustrate that reproducibility is not only paramount to good science in general, but is also a fundamental component of FACT-AI. 
We received very positive feedback on teaching FACT-AI topics through reproducibility. We believe this was an excellent fit for our students, which not only helped motivate them for the duration of the course, but also helped them develop skills that will be essential in their future research careers, whether in the private or public sector. 
To generalize this course setup to other scientific domains, we suggest identifying where the lack of reproducibility in this domain area is coming from and centre the project around evaluating this component.

\begin{table}[htbp!]
\caption{Feedback from students (a) and the MLRC (b, c).}
\centering
\begin{tabular}{@{}l@{}}
\toprule
(a) Feedback from students \\
\midrule
\begin{minipage}[t]{\columnwidth}
\begin{itemize}[leftmargin=*]
    \item ``Reproducing an article was hard and intensive but a really good experience.''
     \item ``Replicating another study, seeing how (poorly) other research is performed was really eye-opening.''
    \item ``Reproducing a paper: I believe this is a good thing to do and is an important part of academia.''
    \item  ``Gave good insights into the trustworthiness of research papers, which is apparently not great.''
    \item ``I appreciate the critical view I have developed on papers as a result of this course. Normally I would easily accept the content of a paper, but I will be more critical from now on, as many papers are not reproducible.''
    \item ``I think it's really good that we get some practical insights into reproducing results from other papers, not all papers are as good as they seem to be.''
    \item ``I really appreciated that this was the first course where students are judging state-of-the-art AI-models. In other words, students were able to experience the scientific workfield of AI.''
\end{itemize}
\end{minipage}
\\
\midrule
(b) Positive feedback from the MLRC \\
\midrule
\begin{minipage}[t]{\columnwidth}
\begin{itemize}[leftmargin=*]
\item ``The report reveals a lot of dark spots of the original paper.''
\item ``Good reviews, strong reproducibility report, provides code reimplementation from scratch which is a strong contribution.''
\item ``The discussion section is a great reference point for future work.''
\item ``The additional experimentation is rather impressive and the report reflects  an intuitive understanding of concepts such as  coverage, correctness, and counterfactual explanations.''
\item ``The report provides good insights on how the experiments in the original paper actually work, while also generating new hypothesis to be tested for future research, which is a positive outcome.''
\end{itemize}
\end{minipage}
\\
\midrule
(c) Negative feedback from the MLRC \\
\midrule
\begin{minipage}[t]{\columnwidth}
\begin{itemize}[leftmargin=*]
\item ``My main concern is that it remains unclear why some of the results are so far off from the original paper?I would have expected the authors to dig deeper on that.''
\item ``The paper is generally difficult to follow. The paper reads closer to an outline than a finished report.''
\item ``It doesn't go above and beyond the reproduction and does not offer novel insights into the workings of the original paper.''
\item ``The submission failed at reproducing the original results. It is unclear whether this is due to a difference in the experimental setup or due to implementation errors. ''
\item ``The paper reads closer to an outline than a finished report. I would encourage the authors to spend some additional time on organization, making sure that the key takeaways are made plain and that the report reads fluidly throughout.''
\end{itemize}
\end{minipage}
\\
\bottomrule
\end{tabular}
\label{tab:feedback}
\end{table}

\section*{Acknowledgements}
We want to thank 
Andreas Panteli, 
Angelo de Groot, 
Christina Winkler, 
Christos Athanasiadis, 
Leon Lang, 
Maartje ter Hoeve, 
Marco Heuvelman, 
Micha de Groot,
Michael Neely, 
Morris Frank, 
Phillip Lippe, 
Simon Passenheim, and
Stefan Schouten 
for their help with the course. 

This research was supported by Ahold Delhaize and the Netherlands Organisation for Scientific Research under project nr.\ 652.\-001.\-003, the Nationale Politie, NWO Innovational Research Incentives Scheme Vidi (016.Vidi.189.039), the NWO Smart Culture - Big Data / Digital Humanities (314-99-301), the H2020-EU.3.4. - SOCIETAL CHALLENGES - Smart, Green And Integrated Transport (814961), and the Hybrid Intelligence Center, a 10-year program funded by the Dutch Ministry of Education, Culture and Science through the Netherlands Organisation for Scientific Research, 
\url{https://hybrid-intelligence-centre.nl}.

All content represents the opinion of the authors, which is not necessarily shared or endorsed by their respective employers and/or sponsors.


\bibliography{main}

\end{document}